\documentclass{article}

\usepackage{PRIMEarxiv}

\usepackage[utf8]{inputenc} 
\usepackage[T1]{fontenc}    
\usepackage{hyperref}       
\usepackage{url}            
\usepackage{booktabs}       
\usepackage{amsfonts}       
\usepackage{nicefrac}       
\usepackage{microtype}      
\usepackage{lipsum}
\usepackage{fancyhdr}       
\usepackage{graphicx}       
\graphicspath{{media/}}     
\usepackage{listings}
\usepackage{xcolor}

\definecolor{codegray}{gray}{0.9}

\lstdefinelanguage{JSON}{
    basicstyle=\ttfamily\small,
    backgroundcolor=\color{codegray},
    keywordstyle=\color{blue},
    stringstyle=\color{black},
    morestring=[b]", 
    morekeywords={Package, attributes, constraints, requirements},
    sensitive=false,
    breaklines=True,
    showstringspaces=false
}

\lstdefinelanguage{SysML}{
    basicstyle=\ttfamily\small,
    backgroundcolor=\color{codegray},
    keywordstyle=\color{blue},
    stringstyle=\color{black},
    morestring=[b]", 
    morekeywords={\%, \{, \}, *, /},
    sensitive=false,
    breaklines=True,
    showstringspaces=false
}

\newcommand{\ST}{\texttt{SysTemp}}
\newcommand{\TGA}{\texttt{TemplateGeneratorAgent}}
\newcommand{\WA}{\texttt{WriterAgent}}
\newcommand{\PA}{\texttt{ParserAgent}}
\newcommand{\RGA}{\texttt{SpecificationGeneratorAgent}}

\title{SysTemp: A Multi-Agent System for Template-Based Generation of SysML v2
}

\author{
  Yasmine Bouamra, Alexandre Poisson \\
  Siemens Digital Industries Software\\
  \texttt{\{yasmine.bouamra, alexandre.poisson\}@siemens.com} \\
   \And
  Bruno Yun, Frédéric Armetta \\
  University Claude Bernard Lyon 1, \\ 
  Ecole Centrale de Lyon, INSA Lyon,\\ 
  Université Lumière Lyon 2, LIRIS, UMR5205,\\ 
  69622 Villeurbanne, France \\
  \texttt{\{bruno.yun,frederic.armetta\}@univ-lyon1.fr} \\
}

\begin{document}
\maketitle

\begin{abstract}
The automatic generation of SysML v2 models represents a major challenge in the engineering of complex systems, particularly due to the scarcity of learning corpora and complex syntax. 
We present SysTemp, a system aimed at facilitating and improving the creation of SysML v2 models from natural language specifications. It is based on a multi-agent system, including a template generator that structures the generation process. We discuss the advantages and challenges of this system through an evaluation, highlighting its potential to improve the quality of the generations in SysML v2 modeling.
\end{abstract}


\section{Introduction}

Model-Based Systems Engenieering (MBSE) \cite{estefan_survey_2008} is an approach that relies on the use of system models (representations of a system’s components, behaviors, and interactions) for analysis, design, and validation in engineering and systems development. MBSE is increasingly adopted as it provides a more rigorous, comprehensive, and collaborative approach to system development compared to the traditional document-based methods. A key driver for MBSE is the fact that 70\% of a product's lifecycle cost is defined during the conceptual phase. MBSE enables systems engineers to capture system requirements, design, analysis, and verification in an integrated model, rather than relying on disconnected files and documents. This model-centric approach helps identify issues earlier, improve design quality, and streamline the development process - overcoming the limitations of the traditional file-based approach that was leading to a dead end. 

\begin{figure}[h]
    \centering
    \includegraphics[width=1\linewidth]{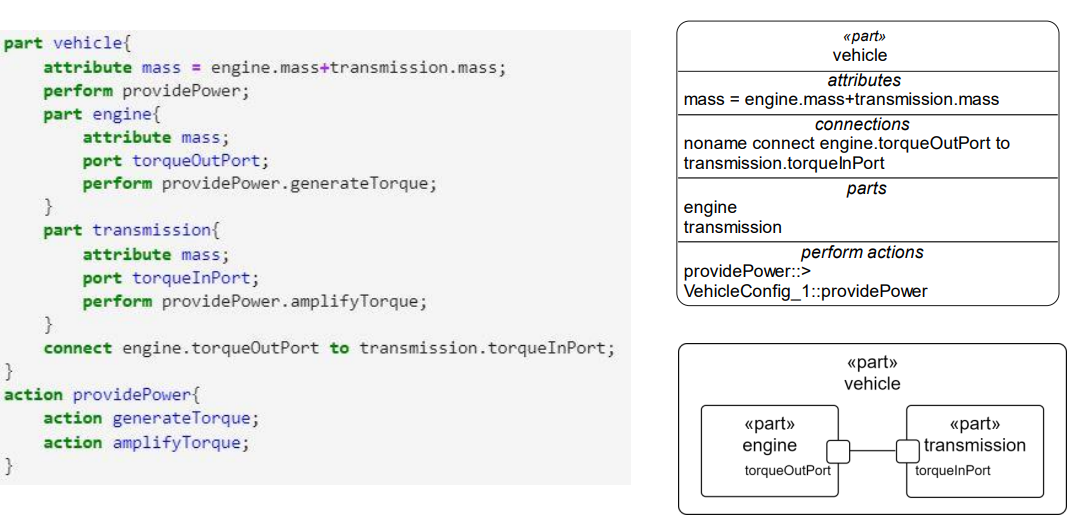}
    \caption{Example of vehicle modeling in SysML v2: The textual notation (left) and the corresponding graphical notation (right) - Source: \url{https://tinyurl.com/ycjbexme} - OMG Slide 18.}
    \label{fig:exemple}
\end{figure}

SysML v1\footnote{See the documentation at \url{https://www.omg.org/spec/SysML/1.0/PDF}.} is the most widely adopted modeling language within MBSE, allowing for the encoding of a wide variety of diagrams. 
It provides a rigorous framework for the detailed description of a system model. However, SysML v1 has limitations, including insufficient expressiveness, increasing complexity in managing large-scale models, and the absence of a standardized textual representation.
To address these challenges, the \textit{Object Management Group (OMG)} has undertaken the development of SysML v2\footnote{The improvements of SysML v2 are presented by Sanford Friedenthal: \url{https://www.omg.org/pdf/SysML-v2-Overview.pdf}.}. 
This second version allows for the translation of specifications expressed in natural language into a formal model, described in a textual notation (see Fig. \ref{fig:exemple} for an illustrative example). 

However, assisting system engineers to write SysML v2 models is complex as (1) the final specification of the language has not been released as its development is still in progress, (2) there are not many available examples (less than 150), and (3) while there is an existing parser\footnote{See \url{https://tinyurl.com/4wauck7j}.}, it does not describe the errors encountered in the model.

Thus, a critical question emerges: \textit{how can we assist system engineers in writing SysML v2 models}?

Recently, several assistant, based on generative artificial intelligence (GenAI), have been integrated into various software environments (such as Visual Studio Code \cite{githubcopilot} and Photoshop \cite{adobe_photoshop_ai}) with the objective of automating repetitive tasks, providing recommendations, and assisting users to reduce their workload. 
Following this trend, we plan to assist users by exploring the challenges associated with translating specifications and instructions formulated in natural language into correct SysML v2 models, leveraging the capabilities of large language models (LLMs). This preliminary work shows promises for future integration into a software assistant dedicated to SysML v2 modeling within \textit{Siemens Digital Industries Software}.

While LLMs have shown excellent performance in user assistance tools (and other downstream tasks like text generation \cite{hao2023boostinglargelanguagemodel}, translation \cite{zhang2025goodllmsliterarytranslation} and code synthesis \cite{jiang2024surveylargelanguagemodels},
due to their ability to capture complex relationships within text (and code), challenges persist in their adaptation to specialized domains with limited or no available data \cite{Huang_2025}. 
Techniques such as few-shot learning \cite{brown2020languagemodelsfewshotlearners} and chain-of-thought \cite{chen2021evaluatinglargelanguagemodels} were developped to tackle these challenges by leveraging LLMs' remarkable generalization capabilities \cite{bender2021dangers}. However, they do not entirely resolve the issue \cite{li2023taskcontaminationlanguagemodels}, highlighting the need for further research into improving LLM adaptability in low-data scenarios.
In the case of SysML v2, initial tests\footnote{See benchmark at \url{https://github.com/yasminebouamra/SysMLv2-Benchmark}.} have reported poor performance of state-of-the-art closed-source (GPT-4 \cite{openai2023gpt4}) and open-source (Llama 3.1-70B \cite{meta2023llama}, Mistral 7B \cite{mistral2023mistral7b}, Gemma\cite{gemmateam2024gemmaopenmodelsbased}) LLM models for related tasks.

To overcome this generalisation problem, we have used a multi-agent approach which has been proven to be particularly suitable for complex tasks \cite{guo2024largelanguagemodelbased}, as it facilitates the decomposition of the task into subtasks and enable successive iterations incorporating feedback and compilation signals.
Our contributions are two-folds: (1) We propose a multi-agent system called \ST, dedicated to SysML v2 code generation, and (2) analyze its effectiveness on several scenarios.

The structure of this paper is as follows. In Section \ref{sota}, we present the background and related work on SysML v2 generation. Section \ref{pipeline} then provides a detailed description of our model and its architecture. The experiments and obtained results are discussed in Sections \ref{eval} and \ref{results}, where we evaluate our approach's performance on a set of representative case studies. Finally, we conclude by summarizing the key contributions of this research and identifying potential improvements and future applications.

\section{Background and Related Work} \label{sota}

\begin{figure}[h]
    \centering
    \includegraphics[width=1\linewidth]{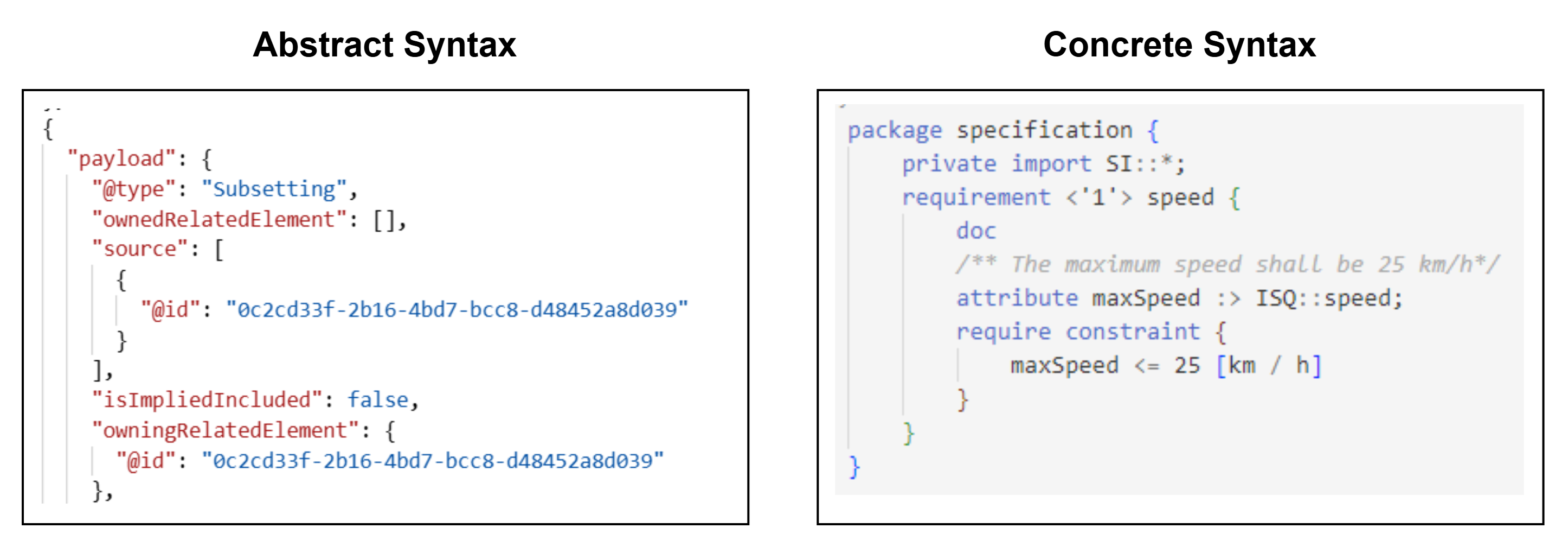}
    \caption{Example of abstract syntax (left) and concrete syntax (right) in SysML v2 -- The abstract syntax is the translation of the concrete syntax but was cropped for readability purposes.}
    \label{fig:syntaxe}
\end{figure}

The automatic generation of SysML v2 models from natural language specifications represents a major challenge in integrating assistance tools for systems engineering. This problem is similar to code generation in that it involves converting a textual description into a formal representation usable by modeling tools. SysML v2 relies on a dual syntax, as shown in Fig. \ref{fig:syntaxe}: an abstract (and machine-oriented) syntax, expressed in JSON, and a concrete syntax, in code form, which is closer to human language and more readable.
It is possible to translate from the abstract to the concrete syntaxe with an existing function.
The abstract syntax is structured as dictionaries of values and is not suited for manual manipulation.  
Automatic handling of SysML v2 is possible through an API\footnote{\url{https://github.com/Systems-Modeling/SysML-v2-API-Services}.} specified by the OMG. It is possible, through API calls, to modify an existing model by adding, and/or removing elements.  
The concrete syntax\footnote{The terms \textit{concrete syntax} and \textit{textual notation} refer to the same thing and are used interchangeably in the article.}, resembles traditional programming languages, it is more concise, and it is more understandable by humans. 
Consequently, we decided to use LLMs to generate the concrete syntax of a model as it is shorter, thus requiring less resources to generate and avoiding  problems with large contexts \cite{an2024doeseffectivecontextlength}.

A major obstacle to applying LLMs for SysML v2 generation lies in the lack of suitable training data.
Unlike traditional programming languages, which have extensive annotated corpora and open-source codebases, SysML v2 resources are limited and heterogeneous\footnote{There are only two available sources: \url{https://github.com/Systems-Modeling/SysML-v2-Pilot-Implementation/tree/master/sysml/src} and \url{https://github.com/GfSE/SysML-v2-Models}, totalling less than 150 SysML v2 scenarios.}. 
From our initial test, even the latest LLMs struggle to generalize effectively for specific modeling tasks in systems engineering, making generation less reliable and more prone to syntactic and semantic errors. 

To date, few studies have combined artificial intelligence with SysML modeling, whether in the initial version or in SysML v2. Among the existing contributions, John K. Dehart's work \cite{dehart2023leveraging} proposes a methodology for the automated generation of SysML v2 API calls to add, remove, or modify model elements. This approach relies on targeted API calls to manipulate specific local properties of the generated model. 
However, their approach requires the user to have an initial SysML v2 model as input. Thus, a promising research avenue would require the exploration of alternative approaches, such as integrating multi-agent systems \cite{vaswani2021multi} capable of combining model generation with iterative verification and correction mechanisms.

The adoption of multi-agent systems for code generation via LLMs has grown significantly in recent years, leading to the development of specialized frameworks such as AutoGen \cite{wu2023autogen}. 
The AutoGen framework allows to orchestrate multiple LLM agents specialized in distinct tasks, such as code generation, syntax validation, and optimization. This methodology is particularly relevant for assisted code generation, where models like PPoCoder \cite{chen2023ppocoder} or StepCoder \cite{dou2024stepcoder} adopt an incremental approach. Instead of producing a program in a single phase, these models decompose the task into sub-problems and iteratively refine the proposed solutions based on interaction feedback. This iterative structuring not only enhances the quality of the generated code but also improves alignment with the initial specifications.

This growing interest in multi-agent orchestration for structured generation tasks extends beyond code to include system modeling. Maria Stella de Biase \cite{debiase2022automaticmodelcompletionrequirements}, for example, explores state machine generation through a refinement process that allows users to iteratively adjust solutions to their specifications.

In this vein, our contribution introduces a multi-agent system augmented by a SysML v2-specific parser and a template-based generator. By leveraging agent collaboration, conversational programming, and user interaction, we aim to enhance automated SysML v2 model generation, producing more accurate system models.

\section{Generative Pipeline Based on Specialist Agents} \label{pipeline}

In this section, we present \ST, a framework designed to guide an LLM in generating SysML v2 models from natural language descriptions.

\begin{figure}
    \centering
    \includegraphics[width=0.8\linewidth]{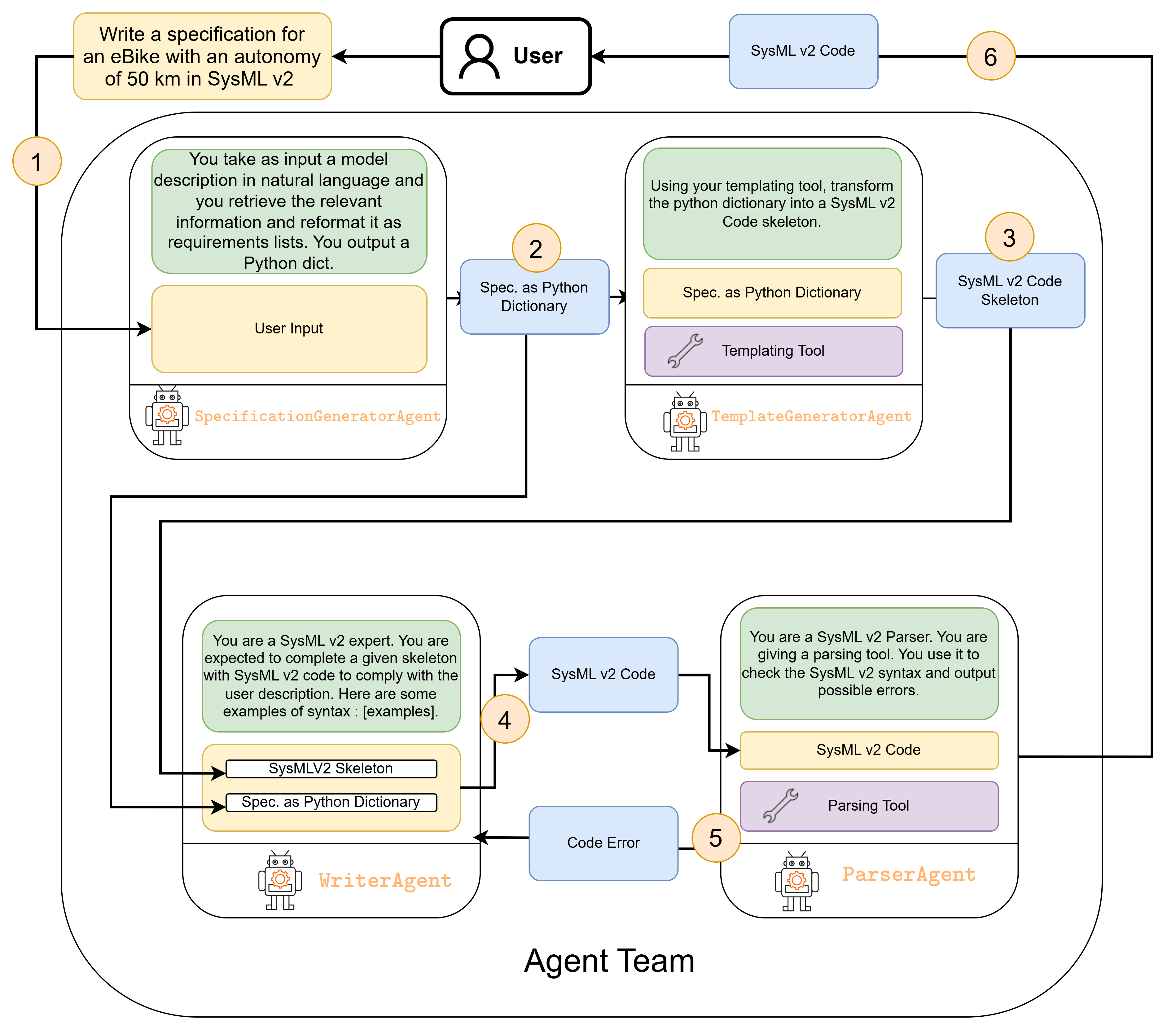}
    \caption{\ST\ pipeline: SysML v2 generation via a multi-agent system.}
    \label{fig:fig-1}
\end{figure}

\ST\ is a multi-agent framework,  illustrated in Fig. \ref{fig:fig-1}, that relies on multiple specialist agents interacting with each other.

We introduce two key specialist agents that characterize our approach: first, the \TGA (TG), which structures the model by generating a syntactically correct SysML v2 skeleton, and second, the \PA, which ensures compliance with specifications by detecting potential syntactic errors (e.g. missing parenthesis, missing semicolumn...)

The \PA\ and \WA\ work together in an iterative loop, where the generated model is progressively refined and adjusted, ensuring conformity to user requirements while minimizing syntactic errors.

Roughly speaking, the generation process unfolds as follows:
\begin{enumerate}
    \item First, the user provides a natural language description of the desired model, detailing system requirements and characteristics.
    \item This input is processed by the \RGA, which extracts key information and reformats it into a structured Python dictionary.
    \item The structured Pytthon dictionary is passed to the \TGA, which generates a syntactically correct SysML v2 skeleton.
    \item The \WA\ uses the skeleton as base and completes it to specify various values (attributes, etc.) consistent with the user's needs, producing a completed SysML v2 model.
    \item The completed model is then analyzed by the \PA, which detects potential syntactic errors or structural inconsistencies and forwards them to the \WA.
    \item Based on this feedback, the \WA\ iteratively adjusts the SysML v2 model until it meets the initial requirements and is free of errors.
\end{enumerate}

In the next section, we formally define the different agents that constitute this pipeline.

\subsection{Pipeline's Agents}
The pipeline consists of four agents, described in the following subsections.
We consider a single LLM as a non-deterministic function $f$ which outputs a string given a string input. This LLM is re-used for all our agents. In our agents, the LLM $f$ may be used to either (1) generate the agent's output directly or (2) provide the input to a tool $t \in \{ \mathcal{P}, \mathcal{T}\} $ that generates the agent's output (described below).

\subsubsection{SpecificationGeneratorAgent}
\label{sec:RGA}
The \RGA\ automatically extracts requirements from a natural language specification. It analyzes the user-provided description to identify key system attributes, structuring them into a Python dictionary for downstream agents.
The \RGA\ employs few-shot prompting. Namely, For a description or query in natural language for the desired model $D_{NL}^q$, the procedure \RGA\ outputs a Python dictionary $Dict^q$ (as a string). An illustrative example of the output is displayed in Fig. \ref{fig:spec-RGA}.
Formally, we have:
\begin{itemize}
    \item \textbf{Input:} A natural language specification $D_{NL}^{q}$,
    \item \textbf{Output:} $Dict^{q} = f(\mathtt{concat}(P_{RGA},Ex,D_{NL}^{q} ))$,
\end{itemize}

    where $P_{RGA}$ is the agent's system prompt\footnote{See Appendix \ref{app:rga}.}, $q$ is the given scenario, $k$ is the number of examples, $Ex = \mathtt{concat}(\{(D_{NL}^{(i)}, Dict^{(i)}) \mid i \in \{ 1, \dots, k\})$ are few-shot examples, $D_{NL}^{(i)}$ is the $i$-th natural language specification, and $Dict^{(i)}$ is the $i$-th corresponding structured Python dictionary.

\begin{figure}[h]
    \centering
    \begin{lstlisting}[language=JSON, frame=single]
        {
            "Package": "BikeFork",
            "attributes": {
                "power": "Rated power in watts (W) or horsepower (HP)",
                ...
            },
            "constraints": {
                ...
            },
            "requirements": {
                "Material": "The bike fork should be made of aluminum.",
                ...
            }
        }
    \end{lstlisting}
    \caption{Specification as Python dictionary obtained from \RGA.}
    \label{fig:spec-RGA}
\end{figure}

\subsubsection{TemplateGeneratorAgent}
\label{sec:TGA}

The \TGA\ is a core component of our approach, responsible for constructing a syntactically correct SysML v2 skeleton (see Fig. \ref{fig:bikefork-spec}) from the structured dictionary generated by the \\ \RGA. This model serves as a foundation for integrating detailed specifications, yielding a format-compliant representation of system requirements.  
\begin{figure}[h]
    \centering
    \begin{lstlisting}[language=SysML, frame=single]
    package BikeFork {
    
        requirement Material 
        {
            doc /* The bike fork should be 
            made of aluminum. */
        }
        
        requirement PivotType 
        {
            doc /* The bike fork should have 
            a 1" 1/8 Aheadset pivot. */
        }
    }
    \end{lstlisting}
    \caption{Skeleton (in textual notation) for a BikeFork SysML v2 Specification.}
    \label{fig:bikefork-spec}
\end{figure}
Formally, we have:

\begin{itemize}
    \item \textbf{Input:} $Dict^{q}$ obtained from \RGA\ (see previous sub-section),
    \item \textbf{Output:} $M^{skeleton}_{SysML} =  \mathcal{T}(f( \mathtt{concat}(P_{TGA}, Dict^q)))$. Here the LLM is used to provide the input to the tool.
\end{itemize}

In the above, $P_{TGA}$ is the agent's system prompt\footnote{See Appendix \ref{app:tga}.}, $\mathcal{T}$ is a tool that applies a sets of rules\footnote{See appendix \ref{app:rules} for full description.} to $Dict^{q}$ by substituting placeholders in a template with corresponding values from $Dict^{q}$. The rules specify the document’s fundamental structure and syntax.
The template tool $\mathcal{T}$ follows an expert system approach leveraging the Jinja2 library\footnote{See \url{https://jinja.palletsprojects.com/en/stable/templates/}.}.
In this agent, the LLM's role is mainly to adapt the input to the tool's signature.

\subsubsection{WriterAgent}  

The \WA\ plays a crucial role in the co-construction process by completing the skeleton generated by the \TGA\ with information from the user’s specifications to obtain a fully specified and SysML v2-compliant model.  
By combining the skeleton with user-provided specific data, this agent ensures model adaptation while maintaining compliance with SysML v2 syntax and principles. Formally, we have:

\begin{itemize}
    \item \textbf{Input:} $M^{skeleton}_{SysML}$, $Dict^{q}$ (see previous sub-sections), and a possible reply $r$ from the \PA\ (see next sub-section).
    \item \textbf{Output:} $ M_{SysML}^{completed}= f( \mathtt{concat}(P_{WA}, Dict^{q}, M^{skeleton}_{SysML}, r))$, 
\end{itemize}
where $P_{WA}$ is the agent's system prompt\footnote{See Appendix, Table \ref{app:wa}.} and $r$ is the empty string initially. 
Note that $P_{WA}$ includes examples illustrating the structure of specific elements such as parts, requirements, and other relevant components. 
The information provided by the \PA\ is then used to iteratively correct syntax until it converges to a specification-compliant version.

\subsubsection{ParserAgent}

The \PA\ is responsible for the syntactic and structural validation of the SysML v2 model. Its primary function is to analyze the model generated by the \WA, identify potential syntax or structural errors, and return a detailed report $E$ specifying the errors and their corresponding locations within the model.  
Formally, we have:

\begin{itemize}
    \item \textbf{Input:} A SysML v2 model \( M_{SysML}^{completed} \) generated by the \WA,
    \item \textbf{Ouput:} $E  = \mathcal{P}(f( \mathtt{concat}(P_{PA}, M_{SysML}^{completed})))$. Here the LLM is used to provide the input to the tool.
\end{itemize}
In the above, $P_{PA}$is the agent's system prompt\footnote{See Appendix, Table \ref{tab:pa_system_prompt}.}, $\mathcal{P}$ is a validation tool that checks whether $M_{SysML}^{completed}$ conforms to the formal grammar defining the SysML v2 syntax, as specified by the OMG SysML v2 standard. 
The parsing mechanism is implemented in Java. 
The identified errors \( E \) are returned to the \WA\, enabling iterative correction until the model fully complies with the SysML v2 specification.  
In this agent, the LLM's role is mainly to adapt the input to the tool's signature.

\section{Evaluation Protocol for the proposed pipeline} \label{eval}

\subsection{Evaluating the Structural Role of the Proposed Model}

The proposed system is evaluated through an ablation study. We aim to measure the effectiveness of the process with and without the \TGA. 
In the ablation setup, the system processes a natural language description without relying on a pre-constructed skeleton.  

Formally, without \TGA, the \WA\ is modified as follows:
\begin{itemize}
    \item \textbf{Input:} $Dict^{q}$ (see previous sub-sections), and a possible reply $r$ from the \PA\ (see next sub-section).
    \item \textbf{Output:} $ M_{SysML}^{completed}= f( \mathtt{concat}(P'_{WA}, Dict, r))$, 
\end{itemize}
where $P'_{WA}$ is the agent's system prompt\footnote{See Appendix, Table \ref{app:wa_wtga}.}. 

The errors are quantified using a counting function that records the number of syntactic errors detected by the \PA.
\subsection{Data \& Models}

The study presented focused on five use cases\footnote{See appendix \ref{app:scenarios} for full description.} described in Appendix. Our examples cover different types of bicycles (mountain bikes, electric bikes, tires, forks, etc.). 
For each scenario, an initial request is written in natural language, and the \TGA\ generates the corresponding skeleton. The scenarios primarily differ in terms of component types. 

For the ablation study, we selected two state-of-the-art LLMs, GPT-4 Turbo and Claude 3.5 Sonnet, due to their strong performance across various downstream tasks.
Additionally, preliminary experiments indicated that open-source LLMs significantly underperform in generating SysML v2 models, making their inclusion less relevant for this study.
Moreover, we used $k = 3$ examples for $\WA$. Literature showed that a good number of examples is needed to infer in few-shot prompting.

\section{Results} \label{results}

\begin{figure}[h]
    \centering
    \includegraphics[width=0.8\linewidth]{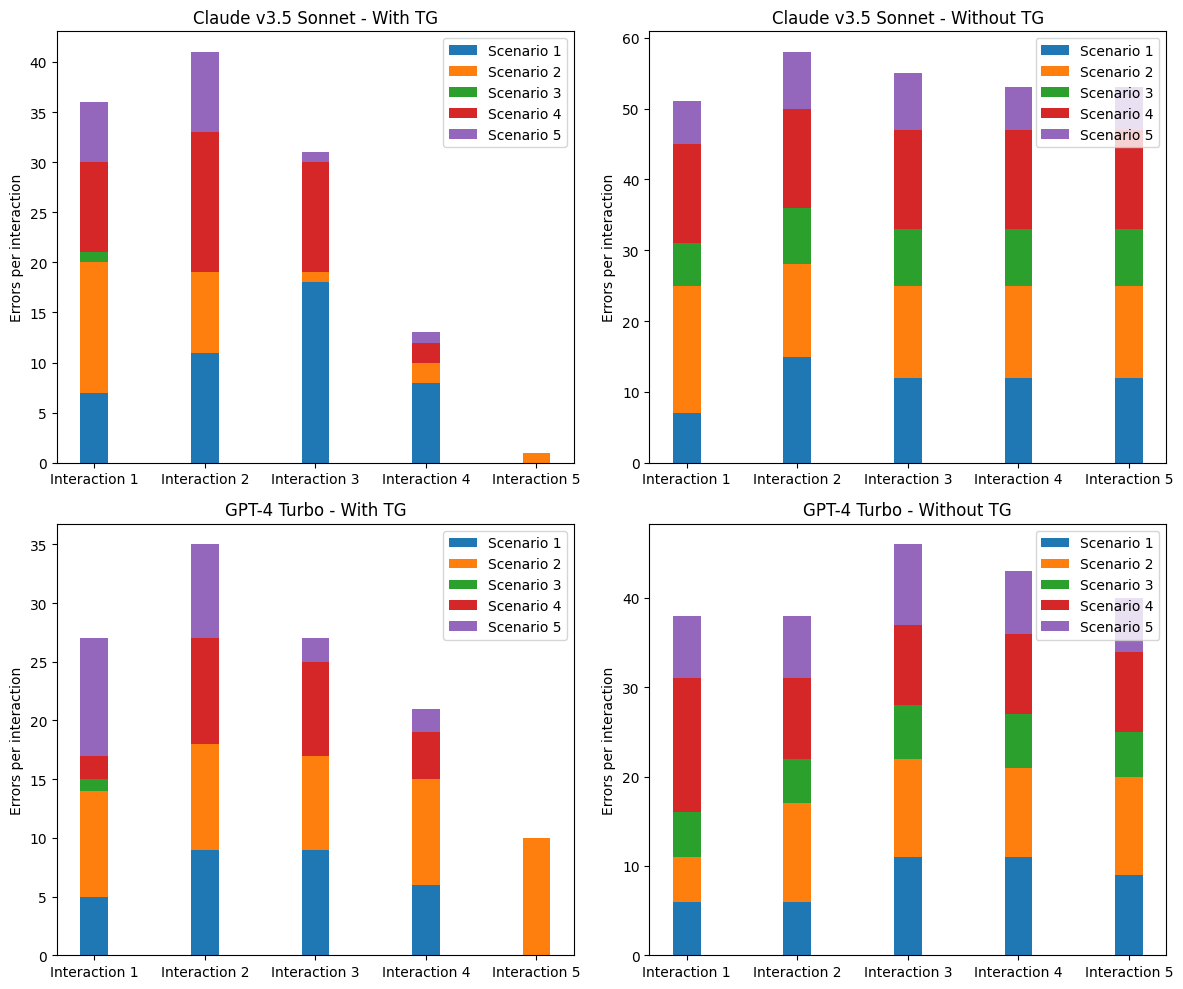}
    \caption{Evolution of the number of errors with and without the \TGA\ for five scenarios.}
    \label{fig:results}
\end{figure}

The results in Fig. \ref{fig:results} show the evolution of syntax errors generated by the LLM during the \WA\ - \PA\ loop. A significant difference is observed between the approaches with and without the \TGA. With the \TGA, near-systematic convergence is achieved, with syntax being correct and validated by the \PA\ in 80\% of the scenarios (4 out of 5). In contrast, without a model-skeleton, convergence (or error-free results at step 5) occurs in only 1 out of 5 scenarios, emphasizing the importance of the initial structuring provided by the skeleton.

Both models (Claude Sonnet 3.5 \cite{anthropic2023claude} and GPT-4 \cite{openai2023gpt4}) yield similar results. A slight reduction in the number of errors is observed with GPT-4. On average, three fewer errors but this difference is not substantial enough to warrant a significant distinction.

\section{Conclusion}
In this paper, we explore the application of LLMs to the automatic generation of SysML v2 models, a modeling language used in systems engineering. Our approach, based on a structured and iterative multi-agent system, investigates the feasibility of using GenAI to address the lack of suitable training data.

We propose \ST, a multi-agent system where we introduce two key agents \TGA\ and \PA, which respectively create a SysML v2 skeleton and provide feedback on possible syntactical errors in the generated SysML v2 code.


Our approach primarily focuses on correcting syntax errors in the model, establishing a foundation upon which further refinements can be built. For instance, future research could explore the semantic quality of generated values. This could be achieved through an iterative generative process where the user specifies the desired characteristics of their design.

However, several challenges remain. On the one hand, the semantics of the generated models require further investigation to ensure the correct interpretation of concepts and optimal alignment with user needs. On the other hand, assessing the quality of generated SysML v2 models remains constrained by the absence of standard benchmarks, highlighting the need to develop specific evaluation metrics tailored to this task. We propose investigating the integration of an LLM-based jury, where multiple instances of the same LLMs are prompted and the most reccurent output is chosen, to remedy these problems. Additionally, incorporating expert knowledge in the form of ontologies or knowledge graphs could enhance the coherence and relevance of the generated models. Our pipeline could also be used for synthetic data generation. Finally, extending this approach to other under-documented modeling languages could further broaden the applicability of LLMs in systems engineering.

Thus, our work lays the foundation for a new methodology for assisted SysML v2 generation and paves the way for future developments aimed at strengthening collaboration between artificial intelligence and systems engineers in advanced modeling contexts.
%
%
%
%

\bibliographystyle{unsrt}  
\bibliography{bib}

\newpage
\appendix
\section{Agents prompts}
\subsection{\RGA} \label{app:rga}
\begin{table}[h]
    \centering
    \begin{tabular}{|p{12cm}|} 
    \hline
    \textbf{System Prompt} \\
    \hline
    You are an extractor-generator agent. You take a description in natural language. You return a python dictionary of packages. The format should be : \\
    
    \{ \\
    \quad "Package" : "PackageName", \\
    \quad "attributes" : \{ \\
    \quad \quad "attribute1" : "attributeName", ... \\
    \quad \}, \\
    \quad "constraints" : \{ \\
    \quad \quad "constraint1" : "constraintName", ... \\
    \quad \}, \\
    \quad "requirements" : \{ \\
    \quad \quad "requirement1": "requirement", ... \\
    \quad \} \\
    \} \\
    
    Each requirement should belong to a package. Each package should have a relevant name. Each requirement should have a description. \\

    \\
    \#\#\#\{Example Description 1 : Example Dict 1\} \#\#\#\\
    \ \ \ \vdots \\
    \#\#\#\{Example Description $k$ : Example Dict $k$\} \#\#\#\\
    \hline
    \end{tabular}
    \caption{\RGA\ system prompt and few-shot examples.}
\end{table}

\subsection{TemplateGeneratorAgent} \label{app:tga}
\begin{table}[h]
    \centering
    \begin{tabular}{|p{12cm}|} 
    \hline
    \textbf{System Prompt} \\
    \hline
    You are a SysML V2 template generator. You take a list of requirements and generate a SysML V2 template containing the requirements. You never return a JSON string. You write only the textual SysML V2 code that you encapsulate between \texttt{''' and '''}. \\

    Example of SysML V2 code is: \\

    \texttt{
    package package\_name \{
        part part\_name \{
            attribute attribute\_name;
        \}
    \}
    } \\
    \hline
    \end{tabular}
    \caption{\TGA\ system prompt.}
    \label{tab:tga_prompt}
\end{table}

\newpage
\subsection{WriterAgent} 

\begin{table}[h]
    \centering
    \begin{tabular}{|p{12cm}|} 
    \hline
    \textbf{System Prompt} \\
    \hline
    You are a SysML V2 code generator. You use your knowledge of SysML V2 to complete a template with a list of requirements. You take a list of requirements and a SysML V2 template as input. You should never add or remove requirements. You should never change the template's structure, only complete it with parts, attributes, constraints, actions, and other relevant elements. You write only the textual SysML V2 code that you encapsulate between \texttt{''' and '''}. \\

    Example of SysML V2 code: \\
    
    \texttt{
    package package\_name \{
        part part\_name \{
            attribute attribute\_name;
        \}
    \}
    } \\

    Another example of SysML V2 code: \\

    \texttt{
    package package\_name \{
        part part\_1;
        part part\_2 \{
            attribute attribute\_name;
        \}
    \}
    } \\

    You ask the \texttt{syntax\_checker\_agent} to check the syntax using only the function provided. \\
    \hline
    \end{tabular}
    \caption{\WA\ system prompt (for use with \TGA).}
    \label{app:wa}
\end{table}

\begin{table}[h]
    \centering
    \begin{tabular}{|p{12cm}|} 
    \hline
    \textbf{System Prompt} \\
    \hline
    You are a SysML V2 code generator. You use your knowledge of SysML V2 to generate a SysML v2 code from a list of requirements. 
    You take a list of requirements as input. 
    You should never add or remove requirements. 
    
    You write only the textual SysML V2 code that you encapsulate between \texttt{''' and '''}. \\

    Example of SysML V2 code: \\
    
    \texttt{
    package package\_name \{
        part part\_name \{
            attribute attribute\_name;
        \}
    \}
    } \\

    Another example of SysML V2 code: \\

    \texttt{
    package package\_name \{
        part part\_1;
        part part\_2 \{
            attribute attribute\_name;
        \}
    \}
    } \\

    You ask the \texttt{syntax\_checker\_agent} to check the syntax using only the function provided. \\
    \hline
    \end{tabular}
    \caption{\WA\ system prompt (for use without \TGA).}
    \label{app:wa_wtga}
\end{table}
\newpage
\subsection{ParserAgent} \label{app:pa}
\begin{table}[h!]
    \centering
    \begin{tabular}{|p{12cm}|} 
    \hline
    \textbf{System Prompt} \\
    \hline
    You are a SysML V2 code parser. You use your knowledge of SysML V2 to check the syntax of the textual code provided by the user. You never return a JSON string. You write only the textual output that you encapsulate between \texttt{''' and '''}. \\

    Example of output: \\

    \texttt{
    '''
    the SysML V2 code contains no error
    '''
    } \\

    Another example of output: \\

    \texttt{
    '''
    Your code contains error:
    Error: Unexpected token 'alias'
    '''
    } \\

    You use the provided function to check the syntax of the code. You do not use any other function. \\
    \hline
    \end{tabular}
    \caption{\PA\ system prompt.}
    \label{tab:pa_system_prompt}
\end{table}

\section{Templating Rules}
\label{app:rules}
The template organizes requirements into \textbf{packages}, where each package groups related requirements. The general structure is:

\begin{lstlisting}[language=SysML, frame=single]
package <package_name> {
    doc /* This is the package containing the requirements */
    requirement <requirement_name> {
        doc /* <requirement_description> */

        attribute <attribute_name> = <value> <units>;

        require constraint {
            <constraint_formula>
        }
    }
}
\end{lstlisting}
The specific rules are defined as follows:

\begin{itemize}
    \item \textbf{Package Definition:} Each package is defined using:
    \begin{lstlisting}[language=SysML, frame=single]
    package {{ package_name }} {
        doc /* This is the package containing the requirements */
    \end{lstlisting}

    \item \textbf{Requirement Definition:} Each requirement inside a package is declared using:
    \begin{lstlisting}[language=SysML, frame=single]
    requirement {{ req.name }} {
        doc /* {{ req.Description }} */
    \end{lstlisting}

    \item \textbf{Attributes (Optional):} If a requirement contains attributes, they are included as:
    \begin{lstlisting}[language=SysML, frame=single]
    {%- for attribute in req.Attributes %}
    attribute {{ attribute.Attribute }} = {{ attribute.Value }} {{ attribute.Units }};
    {%- endfor %}
    \end{lstlisting}

    \item \textbf{Looping Through Packages and Requirements:} The template iterates over all packages and their corresponding requirements using:
    \begin{lstlisting}[language=SysML, frame=single]
    {%- for package_name, reqs in packages.items() %}
    {%- for req in reqs %}
    \end{lstlisting}

\end{itemize}
The final output is stripped of trailing whitespace to ensure a clean SysML v2 code format.

\section{Scenarios}
\label{app:scenarios}
Each scenario follows a structured format:

\begin{itemize}
    \item \textbf{Type:} Indicates that the entry is an input request.
    \item \textbf{Content:} Specifies the requirements for a bicycle or its components in natural language.
\end{itemize}

\begin{enumerate}
    \item \textbf{Mountain Bike Specification}
    \begin{lstlisting}[language=JSON, frame=single]
    {
        "type": "input",
        "content": "Write me a specification for a mountain bike that has:
    
        - An aluminum frame that weighs less than 3 kg.
    
        - A frame with a pronounced sloping design.
    
        - A cassette with 9 cogs ranging from 11 to 42 teeth.
    
        - An aluminum handlebar with a width of 720 mm."
    }
    \end{lstlisting}

    \item \textbf{Electric Bike Specification}
    \begin{lstlisting}[language=JSON, frame=single]
    {
        "type": "input",
        "content": "Write me a specification for an electric bike that has:
    
        - An aluminum hardtail frame suitable for light off-road use.
    
        - 27.5-inch wheels.
    
        - A 380Wh lithium battery."
    }
    \end{lstlisting}

    \item \textbf{Tire Specification}
    \begin{lstlisting}[language=JSON, frame=single]
    {
        "type": "input",
        "content": "Write a specification for tires that must be knobby, sized 24x1.95, 
        and support pressures between 2 and 3.5 bars."
    }
    \end{lstlisting}

    \item \textbf{Mountain Bike with Specific Drivetrain}
    \begin{lstlisting}[language=JSON, frame=single]
    {
        "type": "input",
        "content": "Write me a specification for a mountain bike that meets the following requirements:
    
        - The front suspension must have 50 mm of travel to absorb terrain irregularities.
    
        - The drivetrain must have 7 speeds with an easy trigger shifter.
    
        - The rear derailleur must support a 14/34 freewheel.
    
        - The crankset must have 34 teeth and 152 mm crank arms."
    }
    \end{lstlisting}

    \item \textbf{Bicycle Fork Specification}
    \begin{lstlisting}[language=JSON, frame=single]
    {
        "type": "input",
        "content": """Write me a specification for a bike fork made of aluminum with a 1\" 1/8 Aheadset pivot."
    }
    \end{lstlisting}
\end{enumerate}

\end{document}